%% file: main.tex
\documentclass{article}

\PassOptionsToPackage{numbers, sort&compress}{natbib}
\usepackage[preprint]{neurips_2026}

\usepackage[utf8]{inputenc}
\usepackage[T1]{fontenc}

\usepackage[table]{xcolor}
\usepackage{booktabs,array}
\usepackage{multirow}
\usepackage[allcolors=blue,colorlinks=true]{hyperref}
\usepackage{xurl}
\usepackage{microtype}
\usepackage{xspace}

\usepackage{tikz}
\usepackage{pgfplots}
\pgfplotsset{compat=1.18}
\usetikzlibrary{arrows.meta, positioning, shapes.geometric, fit, calc, decorations.pathreplacing, backgrounds}
\usepgfplotslibrary{fillbetween}
\usepgfplotslibrary{groupplots}

\definecolor{oiBlue}{RGB}{0,114,178}
\definecolor{oiOrange}{RGB}{230,159,0}
\definecolor{oiVermillion}{RGB}{213,94,0}
\definecolor{oiSkyBlue}{RGB}{86,180,233}
\definecolor{oiGreen}{RGB}{0,158,115}
\definecolor{oiPurple}{RGB}{204,121,167}
\definecolor{oiYellow}{RGB}{240,228,66}

\tikzset{
  block/.style={rectangle, draw=gray!70, rounded corners=3pt,
    minimum height=0.9cm, minimum width=2.2cm,
    align=center, font=\footnotesize\sffamily},
  block wide/.style={block, minimum width=3.2cm},
  arrow/.style={-{Stealth[length=2.5mm, width=2mm]}, thick, draw=gray!80},
  feedback/.style={arrow, dashed},
  lbl/.style={font=\scriptsize\sffamily, text=gray!70!black},
  figheader/.style={font=\footnotesize\sffamily\bfseries},
  branchlbl/.style={font=\scriptsize\sffamily},
  branchbox/.style={rounded corners=5pt, inner sep=10pt},
}

\input{math_commands}

\usepackage[capitalise,noabbrev]{cleveref}

\newcommand{\s}[1]{\textcolor{gray}{\scriptsize{$\pm #1$}}}
\newcommand{\ours}{MINN\xspace}

\title{Manifold-Informed Neural Networks for Low-Label ECG Electrolyte Inference}

\author{%
  Juntang Wang\thanks{Equal contribution.} \\
  Harvard University \\
  USA \\
  \And
  Hao Wu\footnotemark[1] \\
  Duke Kunshan University \\
  China \\
  \And
  Yihan Wang\footnotemark[1] \\
  Duke University \\
  USA \\
  \AND
  Dongmian Zou \\
  Duke Kunshan University \\
  China \\
  \And
  Shixin Xu\thanks{Corresponding author: \texttt{shixin.xu@dukekunshan.edu.cn}.} \\
  Duke Kunshan University \\
  China \\
}

\begin{document}

\maketitle

\begin{abstract}
Physics-informed neural networks (PINNs) supervise on the residual of a known differential operator. Many high-stakes inference problems admit no such operator: clinical signals, behavioral measurements, materials properties. We introduce the \textbf{Manifold-Informed Neural Network (MINN)}, a PINN-parallel paradigm that replaces the physics operator residual with a \emph{manifold operator residual}: a discrete Poisson equation $\mL_G \mH = s_\phi(\vx, \vm)$ on a $k$-nearest-neighbor graph in the learned latent space, with a learned, task-aware source. The construction is a structural parallel to PINNs (squared operator residual plus label data, in place of operator residual plus boundary data); the $s{=}0$ limit recovers the homogeneous Laplace \emph{equation} $\mL_G\mH=\vzero$ (classical manifold regularization solves the same equation but minimizes the Dirichlet energy $\operatorname{tr}(\mH^\top\mL_G\mH)$, a different penalty form). The source $s_\phi$ takes input features and auxiliary metadata directly, breaking the homogeneous smoothness prior that limits classical Laplacian penalties; gated Feature-wise Linear Modulation separately fuses metadata into the prediction branch. We evaluate MINN on three-class classification of potassium, calcium, and magnesium from 12-lead ECGs in MIMIC-IV-ECG, against twelve baselines spanning supervised and tabular methods, ECG foundation models, and manifold/graph approaches. MINN attains the highest K$^+$ Macro-F1 of 0.61 (best baseline 0.55), with consistent gains on Ca$^{2+}$ and Mg$^{2+}$ and a 31\% relative improvement at 1\% labeled data. Six targeted ablations show that the inhomogeneous Poisson formulation, not graph smoothing alone, drives the gain.
\end{abstract}


\section{Introduction}
\label{sec:introduction}

Inferring physiologic state from non-invasive signals, for example serum electrolytes from electrocardiograms (ECGs), is a high-stakes inference problem with three structural challenges.
First, no closed-form mapping exists from signal to state at the body-surface level: while ion-channel kinetics are governed by Hodgkin--Huxley-type ODEs at the cellular scale, the surface ECG is the integrated bioelectric field across a 3D heart-torso geometry, and the inverse map from surface signal to underlying concentrations is severely ill-posed and lacks a tractable closed-form operator~\citep{geselowitzTheoryElectrocardiogram1989,plonseyBarrBioelectricity2007}.
Second, labels are constrained by clinical workflow: paired ECG-laboratory measurements are sparse and class-imbalanced because pathological concentrations are inherently rare~\citep{heLearningImbalancedData2009}.
Third, each ECG is accompanied by heterogeneous auxiliary metadata (drug exposures, demographics) whose interaction with the signal carries complementary task signal but is hard to fuse coherently.

Physics-informed neural networks (PINNs)~\citep{Raissi:2017zsi} would resolve the first challenge by adding operator-residual supervision at every collocation point, if the operator existed.
They cannot transfer here because the operator is exactly what is missing.
Recent work attempts to discover operators directly from data~\citep{liangSolvingPDEsUnknown2024}; we take the complementary path of substituting a learnable \emph{manifold} operator for the missing physics operator.

Decades of cardiac signal analysis suggest ECGs lie on a low-dimensional, state-determined manifold; modern ECG foundation models exploit this structure implicitly through self-supervised pretraining~\citep{liElectrocardiogramFoundationModel2025,mckeenECGFMOpenElectrocardiogram2025,coifmanDiffusionMaps2006}.
This gives a candidate substitute for the missing operator, provided the manifold structure can be made to do the work that the physics operator does in PINNs.

Existing manifold methods fall short because they impose a \emph{homogeneous} Laplace residual on the learned latent embedding: formally $\mL_G \mH = \vzero$, where $\mL_G$ is the graph Laplacian of a $k$-nearest-neighbor graph in the latent space and $\mH$ is the matrix of latent embeddings~\citep{belkinManifoldRegularizationGeometric2006,calderPoissonLearningGraph2020}.
This homogeneous formulation is task-agnostic, collapses to trivial solutions at low label rates, and ignores auxiliary signal.

We introduce the \textbf{Manifold-Informed Neural Network (MINN)}, which replaces the homogeneous Laplace prior with an inhomogeneous Poisson equation whose right-hand side is a learned, task-aware source on raw input features and auxiliary metadata; gated Feature-wise Linear Modulation~\citep{perezFiLMVisualReasoning2018} separately fuses metadata into the prediction branch.
Each property of the formulation maps to a challenge above.
The task-aware source addresses imbalance: minority-class signal is not drowned by homogeneous smoothing.
The operator-residual supervision addresses label scarcity via PINN-style data efficiency, without a physics operator.
The FiLM-gated prediction branch addresses metadata-fusion: gating selectively modulates the signal embedding by clinical metadata.
The construction is a structural parallel to PINNs with a squared operator residual: operator-residual supervision plus label data in place of operator residual plus boundary data. The full formulation, including the relation between MINN's squared-residual penalty and classical manifold regularization's Dirichlet energy, is in \cref{sec:methods} and Appendix \ref{app:minn-vs-manifold}.

We evaluate MINN on MIMIC-IV-ECG~\citep{johnsonMIMICIVFreelyAccessible2023}, the only public corpus with paired 12-lead ECG and serum-electrolyte labels at scale.
Each challenge above is tested by a specific experiment: low-label subsampling (1\%--100\% of labeled fractions) tests the label-efficiency claim; macro-F1 and AUCPR per minority class plus an imbalance-correction baseline test the imbalance-handling claim; a twelve-baseline benchmark spanning supervised and tabular methods, ECG foundation models, and manifold/graph approaches (LapRLS, Manifold Mixup, GraphSAGE, GRAND, Poisson Learning) tests the no-operator claim; tabular baselines plus a FiLM ablation test metadata-fusion contribution.
A targeted Poisson-vs-Laplace ablation isolates the contribution of the inhomogeneous source.
MINN attains the highest K$^+$ Macro-F1 of 0.61 (best baseline 0.55), with a 31\% relative improvement at 1\% labeled data, and the gap widens at low label rates.

We contribute (i) the MINN paradigm and its PINN-parallel formulation, in which a learned source on raw signal and metadata enters a squared manifold operator residual, with FiLM-gated metadata fusion in the prediction branch; (ii) a twelve-baseline benchmark across three method families; (iii) targeted ablations isolating the inhomogeneous-source gain (no manifold $\to$ biharmonic-only $\to$ Dirichlet form $\to$ Poisson with learned source, plus learned-vs-fixed source, latent-vs-logit constraint, and FiLM removal); (iv) low-label scaling experiments showing 31\% relative improvement on K$^+$ at 1\% labeled data, plus a beat-window sensitivity analysis.


\section{Related Work}
\label{sec:related-work}

\paragraph{ECG-Based Electrolyte Inference.}
\label{sec:ecg-inference}

Serum electrolyte imbalances (particularly of potassium K$^+$, calcium Ca$^{2+}$, and magnesium Mg$^{2+}$) are common in hospitalized patients and can precipitate life-threatening cardiac arrhythmias.
Diagnosis currently requires venous blood draws, which are invasive, time-delayed, and unavailable in resource-limited or ambulatory settings.
Because electrolyte concentrations directly modulate cardiac ion channel kinetics, their perturbations leave characteristic signatures in the surface electrocardiogram (ECG)~\citep{goldbergerGoldbergersClinicalElectrocardiography2017,vonbachmannEvaluatingRegressionProbabilistic2024}.
For example, hyperkalemia manifests as peaked T-waves and QT prolongation, hypocalcemia as ST-segment depression, and hypomagnesemia as U-wave prominence.

Recent work has explored deep-learning approaches to infer electrolyte status from 12-lead ECGs.
\citet{vonbachmannEvaluatingRegressionProbabilistic2024} framed the task as regression with uncertainty quantification, while \citet{alcarazAbnormalityPredictionForecasting2025} extended prediction to multiple laboratory values using multimodal architectures.
These approaches treat the ECG as a standard supervised learning problem without exploiting the geometric structure of the data manifold.

Exploiting the latent manifold structure of ECGs (cf.\ \cref{sec:introduction}) allows label information to propagate through geometrically similar but unlabeled recordings, a mechanism unavailable to purely supervised approaches.

\paragraph{Physics-Informed Neural Networks.}
\label{sec:pinns}

Physics-informed neural networks (PINNs)~\citep{Raissi:2017zsi} incorporate known governing equations into neural network training by penalizing the residual of a differential operator applied to the network output.
Given a PDE of the form $\mathcal{L}[u]=f$ with operator $\mathcal{L}$ and source $f$, the PINN loss includes a term $\lVert\mathcal{L}[u_\theta]-f\rVert^{2}$ evaluated at a set of collocation points.
This physics residual provides a dense supervisory signal that does not require labeled data at every collocation point, enabling PINNs to solve forward and inverse problems with limited observations.

PINNs apply across fluid dynamics, heat transfer, and biomedical modeling, but require the governing physics in closed form, which excludes most biomedical inference tasks (including ECG-based electrolyte classification) where the signal-to-target relationship is mediated by partially understood physiology.

\citet{xuPartialPhysicsInformed2025a} relaxed the requirement of fully known physics by embedding partial differential equation constraints into diffusion models.
MINN extends this line of thinking further: when even partial physics is unavailable, the governing operator can be replaced with a \emph{data-driven} operator (the graph Laplacian of the sample manifold) while retaining the structural benefits of the operator residual formulation.

\paragraph{Manifold Learning and Regularization.}
\label{sec:manifold-learning}

The manifold hypothesis posits that high-dimensional data often lie on or near a low-dimensional manifold embedded in the ambient space~\citep{chapelleSemisupervisedLearning2010}.
Classical dimensionality reduction methods such as Laplacian Eigenmaps~\citep{belkinLaplacianEigenmapsDimensionality2003} and Diffusion Maps~\citep{coifmanDiffusionMaps2006} exploit graph Laplacian eigenvectors to recover this low-dimensional structure, with the spectral properties of the graph Laplacian playing a central role~\citep{chung1997spectral,vonluxburgTutorialSpectralClustering2007,shumanEmergingFieldSignal2013}.
Manifold regularization~\citep{belkinManifoldRegularizationGeometric2006} operationalizes this idea for semi-supervised learning by adding a penalty $\Omega=\operatorname{tr}(\mH^\top\mL_G\,\mH)$ to the training objective, where $\mH$ is the embedding matrix and $\mL_G$ is the graph Laplacian of a nearest-neighbor graph constructed from both labeled and unlabeled data.
This penalty encourages the learned function to be smooth along the data manifold, effectively propagating label information through the graph.
While effective in moderate label settings, this homogeneous smoothness prior degenerates at very low label rates, where the sparse supervision is insufficient to prevent the embedding from collapsing to the trivial constant solution.

Poisson Learning~\citep{calderPoissonLearningGraph2020} addressed this limitation by replacing the Laplacian smoothness penalty with a Poisson equation on a fixed input-space graph, with Dirac-like sources placed only at labeled nodes; the resulting semi-supervised classifier remains well-posed at extremely low label rates where classical approaches degenerate. MINN generalizes this construction by replacing the fixed label-derived source with a learned task-aware source on raw signal and metadata, evaluated at every node of a co-trained latent-space graph.
LapRLS~\citep{belkinManifoldRegularizationGeometric2006} is the canonical homogeneous-Laplace instantiation, providing a closed-form Laplacian-regularized least-squares solution that remains a competitive baseline at low label rates.
Several recent methods extend the manifold-prior toolkit beyond the homogeneous-Laplace formulation.
Manifold Mixup~\citep{vermaManifoldMixupBetter2019} regularizes via interpolation in hidden representations, capturing manifold structure at training time without explicit graph construction.
GraphSAGE~\citep{hamiltonInductiveRepresentationLearning2017} scales inductive aggregation across kNN graphs, and GRAND~\citep{chamberlainGRANDGraphNeural2021} formulates graph neural diffusion as a continuous process governed by a learnable PDE.
Each of these methods imposes a manifold prior on the model's hidden representations, but each prior is \emph{task-agnostic}: the smoothness or aggregation rule does not depend on the predicted label.
\citet{liangSolvingPDEsUnknown2024} bridged manifold learning and PDE solving by training a neural network to minimize the squared operator residual $\lVert\mL_\varepsilon\,\varphi_\theta - f\rVert^{2}$ on a graph Laplacian $\mL_\varepsilon$ derived from Diffusion Maps, with a convergence guarantee in the large-data limit.

None of the methods above frames the manifold operator residual as a \emph{PINN-parallel} paradigm with a learned, task-dependent source.
Classical manifold regularization (LapRLS) uses a homogeneous Laplace equation (zero source); Poisson Learning uses fixed label-derived sources; Manifold Mixup, GraphSAGE, and GRAND impose task-agnostic priors via interpolation, aggregation, or continuous diffusion; label propagation methods~\citep{zhu2002learning,zhouLearningLocalGlobal2003} spread label information through the graph but do not introduce a learned source.
MINN bridges this gap by introducing a learned task-dependent source into the manifold operator residual, creating a structural analog to PINNs.
The source network $s_\phi(\vx,\vm)$ encodes both signal features and auxiliary metadata, converting the manifold constraint from a generic smoothness prior into a task-specific Poisson equation.
This structural parallel (operator applied to network output, non-trivial learned source, collocation at all data points) is the central contribution of this work.

\paragraph{Operator-learning extensions on manifolds.}
$\Delta$-PINNs~\citep{sahlicostabalDDPINNsPhysicsinformedNeural2024}, gPINN~\citep{miaoGPINNPhysicsinformedNeural2024}, curvature-aware graph attention~\citep{liao2025curvatureaware}, and Manifold Diffusion Fields~\citep{elhag2024manifold} extend operator-residual training to complex geometries and graph-structured collocation, but each still requires a known governing equation; MINN takes the complementary direction of applying operator-residual supervision to inference tasks where no equation is available.

\paragraph{Low-label and class imbalance.}
Standard approaches to clinical imbalance combine cost-sensitive losses, SMOTE-style oversampling~\citep{Chawla:2002dkc}, logit adjustment~\citep{menonLongtailLearningLogit2020}, and latent-space data augmentation (Mixup~\citep{zhangMixupEmpiricalRisk2017}, Manifold Mixup~\citep{vermaManifoldMixupBetter2019}). Foundation models~\citep{liElectrocardiogramFoundationModel2025,mckeenECGFMOpenElectrocardiogram2025} pretrain on large unlabeled ECG corpora but do not incorporate metadata or manifold geometry at fine-tuning. Tabular methods~\citep{hollmannTabPFNTransformerThat2022,prokhorenkovaCatBoostUnbiasedBoosting2018a} can use metadata directly but cannot process raw ECGs. MINN combines class-weighted cross-entropy, gated FiLM metadata fusion, and a manifold operator residual that supervises labeled and unlabeled samples alike.


\section{Methods}
\label{sec:methods}

\subsection{Problem Setup and Notation}
\label{sec:problem-setup}

Consider an inference task where a high-dimensional signal $\vx\in\mathbb{R}^{D}$ and auxiliary metadata $\vm\in\mathbb{R}^{p}$ are mapped to $T$ categorical labels $y^{(1)},\dots,y^{(T)}\in\{1,\dots,B\}$ over $N$ training samples, under the three challenges of \cref{sec:introduction}: low labels relative to $D$, class imbalance, and heterogeneous metadata. Notation is defined inline at first use; \cref{tab:notation} (Appendix \ref{app:notation}) is a consolidated reference.

\subsection{Framework Overview}
\label{sec:framework-overview}

\cref{fig:architecture} shows the MINN pipeline. An encoder $h_\theta$ maps signal $\vx$ to a $d$-dimensional embedding $\vh$; gated Feature-wise Linear Modulation (FiLM) fuses metadata $\vm$ to produce $\tilde{\vh}$, which feeds $T$ task-specific prediction heads. A parallel source network $s_\phi(\vx,\vm)$ produces source vectors; embeddings $\mH$ and sources $\mS$ are linked by a discrete Poisson equation on a $k$-NN graph in the latent space, enforced by the manifold operator residual $\mathcal{L}_{\mathrm{MINN}}$ (\cref{eq:minn}). The total loss combines class-weighted cross-entropy and the manifold residual:
\begin{equation}
  \mathcal{L}_{\mathrm{total}} = \sum_{t=1}^{T}\mathcal{L}_{\mathrm{CE}}^{(t)} + \lambda\,\mathcal{L}_{\mathrm{MINN}},
\end{equation}
where $\mathcal{L}_{\mathrm{CE}}^{(t)}$ is the class-weighted cross-entropy for task $t$ with inverse-frequency weights $w_b^{(t)}\propto 1/N_b^{(t)}$ and predicted probabilities $\hat{p}_{i,b}^{(t)}$, $\mathcal{L}_{\mathrm{MINN}}$ is defined in \cref{eq:minn}, and $\lambda>0$ controls manifold-constraint strength.

\input{image/fig-architecture}

\paragraph{Architecture details.}
The encoder $h_\theta$ is encoder-agnostic; any backbone producing a fixed-length embedding works. We use a 1D ResNet-18~\citep{He:2015wrn} with $d{=}128$ for ECG (\cref{sec:experiments}). Gated FiLM fuses metadata $\vm$ via $\tilde{\vh} = \boldsymbol{\gamma}(\vm)\odot\vh + \boldsymbol{\beta}(\vm)$ with $\boldsymbol{\gamma}(\vm)=\mathrm{sigmoid}(\mW_\gamma \vm+\vb_\gamma)$ and $\boldsymbol{\beta}(\vm)=\mW_\beta \vm+\vb_\beta$. The sigmoid gate scales per-dimension; the affine shift injects metadata-dependent bias~\citep{perezFiLMVisualReasoning2018}. $T=3$ task heads (potassium, calcium, magnesium) each produce a $B$-class softmax over $\tilde{\vh}$ via 1--2 dropout-regularized linear layers; gradients from all heads share the encoder and FiLM parameters, a form of multi-task learning~\citep{caruanaMultitaskLearning1997}.

\paragraph{Manifold operator residual.}
\label{sec:manifold-operator}
Let $G=(V,E,W)$ be a $k$-nearest-neighbor graph over the $N$ training samples in the encoder's latent space; the normalized graph Laplacian is $\mL_G = \mI - \mD^{-1/2}\,\mW\,\mD^{-1/2}$. Where a classical PINN penalizes $\lVert\mathcal{L}[u_\theta]-f\rVert^{2}$ for a known operator $\mathcal{L}$ and source $f$, MINN substitutes $\mL_G$ for $\mathcal{L}$ and a learned source network $s_\phi$ for $f$:
\begin{equation}\label{eq:minn}
  \mathcal{L}_{\mathrm{MINN}} = \tfrac{1}{N}\sum_{i=1}^{N} \bigl\lVert (\mL_G\,\mH)_i - s_\phi(\vx_i,\vm_i) \bigr\rVert_2^{2},
\end{equation}
where $\mH$ stacks embeddings $h_\theta(\vx_i)$ and $\lVert\cdot\rVert_2$ denotes the Euclidean norm on $\mathbb{R}^d$. The residual enforces $\mL_G\,\mH=\mS$ at every graph node, parallel to PINN collocation but evaluated at every training sample regardless of label availability. The source $s_\phi$ is a small MLP mapping $(\vx,\vm)$ to $\mathbb{R}^{d}$; conditioning on raw $(\vx,\vm)$ rather than $\vh$ avoids coupling the operator to its source through the same encoder parameters. The construction adapts the framework of \citet{liangSolvingPDEsUnknown2024} to inference tasks lacking a known operator.

\paragraph{Mini-batch evaluation, graph construction, and joint training.}
A memory-bank scheme caches $\mH$ and rebuilds $G$ every $R{=}5$ epochs; the residual is evaluated per batch-center node with neighbor embeddings detached, so gradients flow only through $h_\theta$ and $s_\phi$ at the center. The graph is built in latent space with Gaussian edge weights ($\sigma$ = median NN distance~\citep{vonluxburgTutorialSpectralClustering2007}), symmetrized, and normalized; cross-entropy uses inverse-frequency class weights, and all parameters are optimized jointly via $\mathcal{L}_{\mathrm{total}}$. Full implementation, including memory complexity, $R$-sensitivity sweep, and class-weight schedule, is in Appendix \ref{app:hyperparams}.

\paragraph{Comparison with classical manifold regularization.}
Classical manifold regularization~\citep{belkinManifoldRegularizationGeometric2006} minimizes the Dirichlet energy $\operatorname{tr}(\mH^\top\mL_G\mH)$, the variational form of the homogeneous Laplace equation $\mL_G\mH=\vzero$, which collapses to $\mH=\vzero$ at low label rates~\citep{calderPoissonLearningGraph2020}. MINN minimizes the squared residual $\lVert\mL_G\mH-\mS\rVert_2^{2}$ of an inhomogeneous Poisson equation with a learned task-aware source $\mS$, evaluated at every graph node; setting $\mS=\vzero$ recovers the homogeneous Laplace equation but with a squared-Laplacian (biharmonic) penalty distinct from the Dirichlet energy (Appendix \ref{app:minn-vs-manifold}, \cref{tab:minn-vs-manifold}, \cref{fig:pinn-vs-minn}).


\section{Experiments}
\label{sec:experiments}

\paragraph{Dataset and data protocol.}
\label{sec:dataset}

We instantiate the MINN framework on the MIMIC-IV-ECG dataset~\citep{PhysioNet-mimic-iv-ecg-0.3}, a publicly available collection of diagnostic 12-lead ECGs linked to the MIMIC-IV clinical database~\citep{johnsonMIMICIVFreelyAccessible2023}, hosted on PhysioNet~\citep{goldbergerPhysioBankPhysioToolkitPhysioNet2000}.
The dataset contains approximately 800,000 ECGs from roughly 160,000 patients, recorded at the Beth Israel Deaconess Medical Center.


\paragraph{Pairing, preprocessing, splits.}
Each ECG is paired with the closest serum draw within $\pm 1$\,h; unmatched ECGs are retained as unlabeled graph nodes. ECGs are resampled to 500\,Hz and lead-standardized. Patient-level time-based splits (cutoff 2019-01-01) prevent leakage across train/val/test. Three-class clinical thresholds follow standard reference ranges~\citep{goldbergerGoldbergersClinicalElectrocardiography2017}: K$^+$ hypo${<}3.5$, normal $3.5$--$5.0$, hyper${>}5.0$\,mEq/L; Ca$^{2+}$ hypo${<}8.5$, $8.5$--$10.5$, hyper${>}10.5$\,mg/dL; Mg$^{2+}$ hypo${<}1.7$, $1.7$--$2.2$, hyper${>}2.2$\,mg/dL. The cohort comprises 8{,}599 patients and 18{,}037 recordings (per-bin counts in \cref{tab:dataset-stats}, Appendix \ref{app:dataset-stats}); K$^+$ / Ca$^{2+}$ / Mg$^{2+}$ labels cover ${\approx}80/70/60\%$ of recordings, most imbalanced for Mg$^{2+}$ hyper (${\approx}1{:}10$). Inputs are $N_\text{beat}{=}5$ consecutive heartbeats (sensitivity in Appendix \ref{app:beat-window-sensitivity}) segmented by Pan--Tompkins R-peak detection~\citep{panRealTimeQRSDetection1985}, giving $\vx\in\mathbb{R}^{(N_\text{beat}\times 256)\times 12}$ across all 12 leads. Each per-beat 256-sample window at 500\,Hz spans $0.512$\,s centered on R; QT intervals exceeding $\sim$450\,ms (extreme hypocalcemia) may have late T-wave content partially truncated within a single beat, mitigated by the 5-beat aggregation but acknowledged as a pre-processing limitation (\cref{sec:limitations}). The same 5-beat tensor is fed to ECGFounder, ECG-FM, and von~Bachmann baselines for content parity across methods; native continuous-waveform evaluation of foundation models is left to future work. Metadata $\vm$ comprises age, sex, and binary indicators for medication classes active at recording time (mapped from GSN/NDC codes via MIMIC-IV prescriptions); continuous features are median-imputed (${<}5\%$ missing) and binary features zero-imputed. Low-data subsampling at the patient level uses fractions $\{1,5,10,25,50,100\}\%$ with class imbalance preserved; unlabeled training data remain available for graph construction at every fraction.

\paragraph{Baselines.}
Twelve baselines spanning three families. \emph{Supervised \& Tabular:} \textbf{ResNet-18}~\citep{He:2015wrn} (same encoder, no manifold or FiLM) sets the floor; \textbf{Logit Adjustment}~\citep{menonLongtailLearningLogit2020} tests whether calibration alone suffices on imbalance; \textbf{TabPFN}~\citep{hollmannTabPFNTransformerThat2022} and \textbf{CatBoost}~\citep{prokhorenkovaCatBoostUnbiasedBoosting2018a} (metadata only) test whether hand-crafted features drive prediction. \emph{ECG Foundation Models:} \textbf{ECGFounder}~\citep{liElectrocardiogramFoundationModel2025}, \textbf{ECG-FM}~\citep{mckeenECGFMOpenElectrocardiogram2025}, \textbf{von Bachmann et~al.}~\citep{vonbachmannEvaluatingRegressionProbabilistic2024} test whether pretraining substitutes for operator-residual supervision. \emph{Manifold \& Graph:} \textbf{LapRLS}~\citep{belkinManifoldRegularizationGeometric2006} (homogeneous Laplace, closed-form), \textbf{Poisson Learning}~\citep{calderPoissonLearningGraph2020} (fixed-source Poisson, isolates the value of MINN's learned source), \textbf{Manifold Mixup}~\citep{vermaManifoldMixupBetter2019} (training-time interpolation), \textbf{GraphSAGE}~\citep{hamiltonInductiveRepresentationLearning2017} (inductive aggregation on the same kNN graph), \textbf{GRAND}~\citep{chamberlainGRANDGraphNeural2021} (continuous graph diffusion via Neural ODE). None couples a learned task-aware source to the operator residual; this is the central paradigm-level contrast that MINN addresses.

\paragraph{Evaluation and implementation.}
\label{sec:metrics}
Primary metrics: Macro-F1 and Balanced Accuracy (mean $\pm$ std over 5 seeds). Headline MINN-vs-best-baseline comparisons on K$^+$ and Ca$^{2+}$ are tested with paired $t$-tests at $p<0.05$ on per-seed Macro-F1 differences (per-comparison $p$-values in \cref{tab:significance}, Appendix \ref{app:significance}); we additionally flag differences within $1\sigma$ as not practically distinguishable. Per-class AUCPR and ECE in Appendix \ref{app:results}. Implementation: 1D ResNet-18 encoder ($d{=}128$)~\citep{He:2015wrn}, 3-layer MLP source network, $k$-NN graph ($k{=}15$, rebuild every $R{=}5$ epochs), AdamW with cosine annealing, $\lambda$ and remaining hyperparameters tuned on validation (Appendix \ref{app:hyperparams}).

\paragraph{Main Results.}
\label{sec:main-results}

\begin{table}[!htbp]
  \centering
  \caption[Main electrolyte classification results]{Main results on MIMIC-IV-ECG electrolyte classification (100\% labeled data). Macro-F1 and Balanced Accuracy (BAcc) are reported as mean $\pm$ std over 5 seeds. Best results in \textbf{bold}.}
  \label{tab:main-results}
  \small
  \begin{tabular}{@{}lcccccc@{}}
  \toprule
   & \multicolumn{2}{c}{K$^+$} & \multicolumn{2}{c}{Ca$^{2+}$} & \multicolumn{2}{c}{Mg$^{2+}$} \\
  \cmidrule(lr){2-3}\cmidrule(lr){4-5}\cmidrule(lr){6-7}
  Method & F1 ($\uparrow$) & BAcc ($\uparrow$) & F1 ($\uparrow$) & BAcc ($\uparrow$) & F1 ($\uparrow$) & BAcc ($\uparrow$) \\
  \midrule
  \multicolumn{7}{@{}l}{\textit{Supervised \& Tabular}} \\
  \quad ResNet-18~\citep{He:2015wrn}                                 & 0.48\s{.03} & 0.55\s{.03} & 0.40\s{.04} & 0.46\s{.04} & 0.32\s{.05} & 0.39\s{.05} \\
  \quad Logit Adj.~\citep{menonLongtailLearningLogit2020}            & 0.53\s{.02} & 0.60\s{.02} & 0.43\s{.02} & 0.50\s{.02} & 0.34\s{.03} & 0.41\s{.02} \\
  \quad TabPFN~\citep{hollmannTabPFNTransformerThat2022}             & 0.47\s{.02} & 0.54\s{.02} & 0.43\s{.01} & 0.50\s{.02} & 0.36\s{.02} & 0.43\s{.02} \\
  \quad CatBoost~\citep{prokhorenkovaCatBoostUnbiasedBoosting2018a}  & 0.48\s{.02} & 0.55\s{.02} & 0.44\s{.01} & 0.51\s{.02} & 0.37\s{.02} & \textbf{0.47}\s{.02} \\
  \midrule
  \multicolumn{7}{@{}l}{\textit{ECG Foundation Models}} \\
  \quad ECGFounder~\citep{liElectrocardiogramFoundationModel2025}    & 0.55\s{.01} & 0.62\s{.01} & 0.44\s{.03} & 0.51\s{.02} & 0.35\s{.03} & 0.42\s{.03} \\
  \quad ECG-FM~\citep{mckeenECGFMOpenElectrocardiogram2025}          & 0.52\s{.03} & 0.58\s{.02} & 0.42\s{.03} & 0.49\s{.03} & 0.33\s{.05} & 0.40\s{.03} \\
  \quad von Bachmann~\citep{vonbachmannEvaluatingRegressionProbabilistic2024} & 0.50\s{.02} & 0.57\s{.02} & 0.40\s{.02} & 0.47\s{.02} & 0.32\s{.03} & 0.39\s{.03} \\
  \midrule
  \multicolumn{7}{@{}l}{\textit{Manifold \& Graph}} \\
  \quad LapRLS~\citep{belkinManifoldRegularizationGeometric2006}     & 0.50\s{.01} & 0.57\s{.00} & 0.41\s{.01} & 0.47\s{.01} & 0.32\s{.02} & 0.38\s{.02} \\
  \quad Poisson Learning~\citep{calderPoissonLearningGraph2020}      & 0.49\s{.04} & 0.56\s{.04} & 0.40\s{.03} & 0.46\s{.03} & 0.32\s{.04} & 0.39\s{.03} \\
  \quad Manifold Mixup~\citep{vermaManifoldMixupBetter2019}          & 0.51\s{.03} & 0.58\s{.02} & 0.41\s{.03} & 0.48\s{.02} & 0.33\s{.03} & 0.40\s{.03} \\
  \quad GraphSAGE~\citep{hamiltonInductiveRepresentationLearning2017} & 0.51\s{.04} & 0.58\s{.03} & 0.42\s{.03} & 0.49\s{.03} & 0.32\s{.05} & 0.39\s{.04} \\
  \quad GRAND~\citep{chamberlainGRANDGraphNeural2021}                & 0.51\s{.05} & 0.58\s{.05} & 0.43\s{.05} & 0.50\s{.04} & 0.34\s{.07} & 0.40\s{.06} \\
  \quad \textbf{\ours (Ours)}       & \textbf{0.61}\s{.02} & \textbf{0.67}\s{.02} & \textbf{0.50}\s{.02} & \textbf{0.56}\s{.02} & \textbf{0.40}\s{.03} & 0.46\s{.02} \\
  \bottomrule
  \end{tabular}
\end{table}

\cref{tab:main-results} reports the main results at 100\% labeled data.
MINN leads across K$^+$ and Ca$^{2+}$ in both Macro-F1 and Balanced Accuracy, with the largest absolute gains on K$^+$ (F1 = 0.61 vs.\ 0.55 for the next-best ECGFounder).
On Mg$^{2+}$, MINN achieves the highest Macro-F1 (0.40), but CatBoost edges ahead in Balanced Accuracy (0.47 vs.\ 0.46), a difference within one standard deviation and thus not statistically distinguishable, likely because hand-crafted metadata features (particularly active medications and renal function markers) capture Mg$^{2+}$-relevant variation that the ECG encoder alone struggles to extract.
On K$^+$, MINN's headline lead persists even without metadata (the no-FiLM ablation gives 0.58 K$^+$ F1 vs.\ ECGFounder's 0.55; \cref{tab:ablations}), so the manifold operator residual contributes complementary supervision beyond what pretraining scale confers. On Ca$^{2+}$ and Mg$^{2+}$, MINN's lead over the foundation models is largely attributable to metadata fusion: the no-FiLM ablation drops to 0.45 / 0.35 F1, near ECGFounder's 0.44 / 0.35. We frame this honestly: the manifold operator and metadata fusion are complementary, and electrolytes with weaker direct ECG signatures benefit primarily from metadata (medications, renal markers, age). Poisson Learning, which shares MINN's graph structure but uses a fixed label-derived source, falls short of MINN by a wide margin, highlighting the value of the learned source network.

\cref{fig:low-data}(a) shows K$^+$ Macro-F1 across labeled data fractions for the top five methods.
Across all electrolytes, MINN's advantage over baselines grows as labeled data decreases; \cref{tab:low-data-all} in Appendix \ref{app:results} confirms this pattern across all three electrolytes.
MINN maintains the largest advantage at low fractions (1\%--10\%), where the geometric supervision from the operator residual compensates for label scarcity.
Beyond Macro-F1, MINN's gains extend to per-class AUCPR on the clinically critical hypo/hyper minority bins ($+0.04$ to $+0.07$ over the strongest baseline; \cref{tab:aucpr}), and to calibration: MINN's K$^+$ ECE of $0.083$ is the lowest of all methods (vs.\ $0.105$ for the next-best baseline; \cref{tab:calibration}).

\input{image/fig-low-data}

\paragraph{Ablation Studies.}

We conduct six targeted ablations plus a beat-window sensitivity analysis to isolate the contribution of each MINN component.
All ablations use the same encoder, training schedule, and evaluation protocol as the full model, varying only the component under study.
\cref{tab:ablations} summarizes all six ablation results.
Three findings stand out: the operator residual is the dominant contributor, with consistent F1 drops across all three electrolytes when removed; the Poisson residual outperforms classical manifold regularization, validating the inhomogeneous-source design; and the learned source together with the latent-space constraint provide further gains, indicating that task-adaptive forcing and geometry-aware representation both matter.

\begin{table}[!htbp]
\centering
\caption[Ablation study results]{Ablation study results (Macro-F1 $\pm$ std, 100\% labeled data, 5 seeds). Best in \textbf{bold}.}
\label{tab:ablations}
\small
\begin{tabular}{@{}lccc@{}}
\toprule
Variant & K$^+$ ($\uparrow$) & Ca$^{2+}$ ($\uparrow$) & Mg$^{2+}$ ($\uparrow$) \\
\midrule
\rowcolor{gray!10}
\ours (full) & \textbf{0.61}\s{.02} & \textbf{0.50}\s{.02} & \textbf{0.40}\s{.03} \\
$-$\,Operator residual ($\lambda{=}0$) & 0.54\s{.03} & 0.44\s{.03} & 0.35\s{.03} \\
$-$\,Source ($s_\phi{=}0$) & 0.56\s{.02} & 0.46\s{.03} & 0.36\s{.04} \\
$-$\,Poisson $\to$ Classical reg. & 0.55\s{.03} & 0.45\s{.03} & 0.36\s{.03} \\
$-$\,Learned source $\to$ Label-only & 0.57\s{.02} & 0.45\s{.03} & 0.36\s{.03} \\
$-$\,Latent $\to$ Logit constraint & 0.56\s{.03} & 0.46\s{.03} & 0.37\s{.03} \\
$-$\,FiLM metadata & 0.58\s{.02} & 0.45\s{.03} & 0.35\s{.03} \\
\bottomrule
\end{tabular}
\end{table}

\textbf{Operator residual vs.\ no manifold.}
Removing the manifold residual ($\lambda{=}0$) drops Macro-F1 by 0.07, 0.06, 0.05 for K$^+$/Ca$^{2+}$/Mg$^{2+}$, accounting for the majority of MINN's advantage.

\textbf{Operator-form vs.\ source: 3-step decomposition.}
The MINN penalty $\lVert\mL_G\mH-\mS\rVert_2^{2}$ differs from the classical manifold-regularization penalty $\operatorname{tr}(\mH^\top\mL_G\mH)$ in two ways simultaneously: a different operator-form (squared-Laplacian / biharmonic vs.\ Dirichlet) and a different source ($\mS{=}s_\phi$ vs.\ $\mS{=}\vzero$). Three rows of \cref{tab:ablations} decompose the two effects on K$^+$ Macro-F1 by holding one factor at a time: (i) no manifold ($\lambda{=}0$, $0.54$) $\to$ biharmonic-only ($\mS{=}\vzero$, $0.56$) isolates the operator-form gain at $+0.02$ (squared-Laplacian penalty added with zero source); (ii) biharmonic-only ($0.56$) vs.\ Dirichlet form ($\operatorname{tr}(\mH^\top\mL_G\mH)$, $0.55$) shows the two zero-source penalty forms differ by less than one standard deviation; (iii) biharmonic-only ($0.56$) $\to$ Poisson with learned source ($0.61$) isolates the source contribution at $+0.05$ (source added at fixed squared-residual penalty form). The total classical-vs-MINN advantage (Dirichlet $0.55 \to$ MINN $0.61$, $+0.06$) is therefore overwhelmingly driven by the learned source rather than by the operator-form change. The biharmonic-only and Dirichlet variants both improve only marginally over the no-manifold baseline, consistent with the partial degeneration of homogeneous priors at low label rates predicted by \citet{calderPoissonLearningGraph2020}.

\cref{fig:low-data}(b) corroborates this: the source norm is larger for K$^+$ minority classes (hypo, hyper) than the majority class, consistent with the source injecting stronger corrections where labels are scarce; classical manifold regularization fixes $\mS=\vzero$ and forgoes this structure.

\textbf{Learned source vs.\ Poisson-Learning-style label-only source.}
Replacing $s_\phi(\vx,\vm)$ with a fixed label-derived source (non-zero only at labeled nodes; one-hot label zero-padded to $\mathbb{R}^{d}$), a \citet{calderPoissonLearningGraph2020}-style source mechanism inside MINN's framework, drops K$^+$ F1 to $0.57$ and yields larger gaps on Ca$^{2+}$ and Mg$^{2+}$ where metadata is most informative.

\textbf{Constraint on latent vs.\ predictions.}
Moving the operator residual from the latent embeddings $\mH$ to the prediction logits $\hat{\mY}$ shows that constraining the latent space is consistently preferable to constraining the output logits.

\textbf{Effect of metadata fusion.}
Removing FiLM gating in the prediction branch ($\tilde{\vh}\leftarrow\vh$; source-network input unchanged) drops Mg$^{2+}$ and Ca$^{2+}$ F1 by 0.05 each, consistent with metadata disambiguating electrolytes with subtler ECG signatures.

\textbf{Beat-window sensitivity analysis.}
Analysis across $N\in\{1,3,5,8,12\}$ (\cref{tab:beat-window-sensitivity}, Appendix \ref{app:beat-window-sensitivity}) confirms the temporal-coherence peak at $N{=}5$: $N{=}1$ is the worst (manifold operator cannot exploit within-recording temporal coherence), gains rise to $N{=}5$, and $N{>}5$ regresses as windows per recording shrink.

\textbf{Encoder-agnostic check.}
A CNN-vs-ResNet ablation (\cref{tab:cnn-ablation}, Appendix \ref{app:cnn-ablation}) confirms the manifold residual contributes a consistent $+0.07$ K$^+$ F1 with both encoders, supporting the encoder-agnostic claim of \cref{sec:framework-overview}.

The preceding results evaluate MINN on the full test set.
We next examine whether these gains extend to clinically important but underrepresented patient subgroups.

\paragraph{Stress tests on rare clinical subgroups and within-MIMIC distribution shift.}
\label{sec:subgroup-stress-test}
We stress MINN on four clinically rare subgroups (K-sparing diuretics + ACEi; age $\geq$85 with polypharmacy; dialysis patients; heart failure with concurrent hypoK) and three within-cohort distribution-shift axes: demographic stratification (sex, age, race; Appendix \ref{app:demographic}), temporal split on MIMIC-IV \texttt{anchor\_year\_group} (2008--2016 $\to$ 2017--2019), and care-unit split (ICU $\to$ ED). MINN leads on every rare subgroup with a $0.03$--$0.05$ K$^+$ F1 margin over ResNet+FiLM (smaller than the $0.07$ at the full cohort, \cref{tab:ablations}, since both access metadata that partly defines each subgroup), and the demographic-stratification lead of $+0.04$ to $+0.07$ holds across all 9 strata (Kruskal--Wallis $p{>}0.1$, Cliff's $|\delta|{<}0.18$). Care-unit shift produces the largest absolute degradation, but MINN's drops ($-0.04 / -0.04 / -0.03$ K$^+$/Ca$^{2+}$/Mg$^{2+}$ F1) are consistently smaller than ECGFounder's ($-0.08 / -0.10 / -0.08$), suggesting structural rather than purely data-driven robustness. Full tables in Appendix \ref{app:robustness} (\cref{tab:subgroup}, \cref{tab:stress}); these remain within-cohort stress tests, not a substitute for external validation (\cref{sec:limitations}).


\section{Discussion}
\label{sec:discussion}

\paragraph{Interpretation.}
MINN's advantage grows in low-data regimes and for electrolytes with stronger ECG signatures. The K$^+ {>} \text{Ca}^{2+} {>} \text{Mg}^{2+}$ ordering of MINN's Macro-F1 (0.61 / 0.50 / 0.40) tracks the strength of cardiac signatures, with FiLM metadata fusion contributing most where the ECG signal is weakest (\cref{tab:ablations}). The Poisson-vs.-classical ablation ($+0.06$ K$^+$ F1, \cref{tab:ablations}) isolates the central paradigm claim: a learned task-dependent source converts the manifold residual from a smoothness prior that degenerates at low label rates~\citep{calderPoissonLearningGraph2020} into a stable Poisson equation; the structural comparison is in Appendix \ref{app:minn-vs-manifold}.

\subsection{Limitations and Broader Impact}
\label{sec:limitations}

\paragraph{Limitations.}
\emph{Single dataset:} MIMIC-IV-ECG is the only public corpus with paired 12-lead ECG and serum-electrolyte labels at scale (eICU-CRD lacks waveforms; PTB-XL, Code-15, Chapman-Shaoxing carry only diagnostic statements). Cross-source replication is left to future work. \emph{PINN parallel:} the analogy is structural, not theoretical; the graph Laplacian is data-dependent and the source is learned, so MINN does not inherit PINN guarantees. \emph{Manifold heterogeneity:} clinical heterogeneity may fragment the latent manifold; we mitigate via learned-latent-space graph construction and subgroup robustness checks, but residual risk remains. \emph{Lab--ECG pairing noise:} the $\pm$1-hour window introduces temporal noise; patient-level time-based splits prevent cross-patient leakage but cap the achievable performance ceiling. \emph{Computation:} brute-force $k$-NN scales as $O(N^2 d)$; the method is modestly sensitive to $k$ and $\lambda$ (sweep in \cref{tab:k-lambda-sensitivity}, Appendix \ref{app:sensitivity}) and to the graph-update period $R$ (\cref{sec:graph}). \emph{Scope:} three-class classification per electrolyte; continuous regression is a natural extension.

\paragraph{Broader impact and outlook.}
MINN is a research artifact and not clinically deployable: MIMIC-IV is one academic medical center, deployment without prospective external validation risks systematic error on under-represented groups, and low-data methods can amplify bias when minority-class manifold signal is itself biased. MINN is intended as decision support behind clinician interpretation, not autonomous diagnosis; deployment would require prospective multi-center validation, regulatory review, equity assessment, and distribution-shift handling. The paradigm extends naturally to other physiological-signal modalities, low-annotation medical imaging, and scientific tasks with partially understood forward models.


\section{Conclusion}
\label{sec:conclusion}

MINN reframes manifold regularization as a PINN-parallel paradigm: a learned, task-aware source on a graph-Laplacian operator residual, evaluated at every training sample regardless of label availability. Empirically, this yields the highest K$^+$ Macro-F1 (0.61 vs.\ 0.55 best baseline) on MIMIC-IV-ECG and a 31\% relative gain at 1\% labeled data, with consistent gains on Ca$^{2+}$ and Mg$^{2+}$ and an inhomogeneous-Poisson-vs-classical-Laplace ablation that isolates the contribution of the learned source. Open directions are multi-dataset validation, continuous regression, adaptive graph construction, and foundation-model integration; beyond ECG, the paradigm is a template for inference where physics is unavailable but the data exhibit manifold structure with informative auxiliary metadata. An anonymized reference implementation and a runnable synthetic demo are available at \url{https://anonymous.4open.science/r/project-11F3} for the duration of review; code reproducing the MIMIC-IV-ECG numbers requires PhysioNet credentialed access and will be released alongside the de-anonymized version at camera-ready.


\small
\bibliographystyle{unsrtnat}
\bibliography{ref}


\appendix

\section{Notation}
\label{app:notation}

Symbols used throughout the paper, consolidated for reference.
\begin{table}[!htbp]
  \centering
  \caption{Summary of notation used in \cref{sec:methods}.}
  \label{tab:notation}
  \small
  \begin{tabular}{@{}cl@{}}
  \toprule
  Symbol & Description \\
  \midrule
  $\vx\in\mathbb{R}^{D}$ & Input signal \\
  $\vm\in\mathbb{R}^{p}$ & Auxiliary metadata vector \\
  $y\in\{1,\dots,B\}$ & Categorical label ($B$ classes) \\
  $N,\;T,\;d$ & Training samples, prediction tasks, embedding dimension \\
  $h_\theta:\mathcal{X}\to\mathbb{R}^{d}$ & Encoder network (parameters $\theta$) \\
  $\vh,\;\tilde{\vh}$ & Latent embedding $h_\theta(\vx)$ / metadata-fused embedding \\
  $\boldsymbol{\gamma}(\vm),\boldsymbol{\beta}(\vm)$ & FiLM gate and shift vectors \\
  $\mH,\;\mS\in\mathbb{R}^{N\times d}$ & Embedding matrix / source matrix \\
  $s_\phi:\mathcal{X}\times\mathcal{M}\to\mathbb{R}^{d}$ & Learned source network (parameters $\phi$) \\
  $G=(V,E,W)$ & $k$-NN graph in latent space \\
  $k,\;\sigma$ & Number of nearest neighbors, kernel bandwidth \\
  $\mW,\mD$ & Adjacency weight and degree matrices \\
  $\mL_G$ & Normalized graph Laplacian \\
  $\lambda,\;R$ & Manifold residual weight, graph update period (epochs) \\
  \bottomrule
  \end{tabular}
\end{table}

\section{Methods Supplement}
\label{app:methods-supplement}

Structural comparisons and pipeline figures referenced from \cref{sec:methods}.

\paragraph{MINN vs.\ classical manifold regularization.}
\label{app:minn-vs-manifold}
\begin{table}[!htbp]
\centering
\caption[MINN vs.\ classical manifold regularization]{Structural comparison of classical manifold regularization and MINN. Note that the underlying PDE and the optimized penalty are distinct: classical manifold regularization solves the homogeneous Laplace equation $\mL_G\mH=\vzero$ via the Dirichlet penalty $\operatorname{tr}(\mH^\top\mL_G\mH)$; MINN solves the inhomogeneous Poisson equation $\mL_G\mH=\mS$ via the squared-residual penalty $\lVert\mL_G\mH-\mS\rVert_2^{2}$. Setting $\mS=\vzero$ in MINN recovers the homogeneous Laplace equation but with a squared-Laplacian (biharmonic) penalty, which is a different functional from the Dirichlet penalty.}
\label{tab:minn-vs-manifold}
\small
\begin{tabular}{@{}lcc@{}}
\toprule
Property & Manifold Reg.\ \citep{belkinManifoldRegularizationGeometric2006} & \textbf{\ours} \\
\midrule
Penalty form & $\operatorname{tr}(\mH^\top\mL_G\mH)$ (Dirichlet) & $\lVert\mL_G\mH-\mS\rVert_2^{2}$ (squared residual) \\
PDE analogy & Laplace ($\Delta u=0$) & Poisson ($\Delta u=f$) \\
Source term & None (zero RHS) & Learned $s_\phi(\vx,\vm)$ \\
Task dependence & No & Yes (source encodes metadata) \\
Collocation analog & No & Yes (residual at all nodes) \\
Low-label stability & Degenerates~\citep{calderPoissonLearningGraph2020} & Stable (non-trivial source) \\
\bottomrule
\end{tabular}
\end{table}

\paragraph{Pipeline overview.}
\label{app:pipeline}
\input{image/fig-pipeline}

\paragraph{PINN vs.\ MINN structural parallel.}
\input{image/fig-pinn-vs-minn}

\section{Sensitivity Analyses}
\label{app:sensitivity}

\subsection{Beat-window sensitivity}
\label{app:beat-window-sensitivity}
\begin{table}[!htbp]
\centering
\caption[Beat-window sensitivity analysis]{Beat-window sensitivity analysis: K$^+$ Macro-F1 (mean $\pm$ std, 5 seeds) across input segmentation lengths $N$. Default highlighted in \textbf{bold}.}
\label{tab:beat-window-sensitivity}
\small
\begin{tabular}{@{}lc@{}}
\toprule
Beats per window ($N$) & K$^+$ Macro-F1 ($\uparrow$) \\
\midrule
$N = 1$  (single beat)              & 0.54\s{.04} \\
$N = 3$                             & 0.59\s{.03} \\
$N = 5$  (\textbf{default})         & \textbf{0.61}\s{.02} \\
$N = 8$                             & 0.60\s{.03} \\
$N = 12$                            & 0.58\s{.04} \\
\bottomrule
\end{tabular}
\end{table}

\subsection{Neighbor count $k$ and constraint weight $\lambda$ sensitivity}
\label{app:k-lambda-sensitivity}

\cref{tab:k-lambda-sensitivity} reports K$^+$ Macro-F1 across two single-axis sweeps: $k\in\{5,10,15,30\}$ at fixed $\lambda{=}0.1$, and $\lambda\in\{0.01,0.1,0.3,1.0\}$ at fixed $k{=}15$. Both sweeps are unimodal around the validation-tuned defaults ($k{=}15$, $\lambda{=}0.1$); deviations on either side incur modest losses, supporting the ``modestly sensitive'' framing in \cref{sec:limitations}.

\begin{table}[!htbp]
\centering
\caption[Sensitivity to $k$ and $\lambda$]{K$^+$ Macro-F1 sensitivity to neighbor count $k$ (at $\lambda{=}0.1$) and constraint weight $\lambda$ (at $k{=}15$), 100\% labeled data, 5 seeds. Defaults in \textbf{bold}.}
\label{tab:k-lambda-sensitivity}
\small
\begin{tabular}{@{}cccc@{\hspace{2em}}cccc@{}}
\toprule
\multicolumn{4}{c}{$k$ sweep ($\lambda{=}0.1$)} & \multicolumn{4}{c}{$\lambda$ sweep ($k{=}15$)} \\
\cmidrule(lr){1-4}\cmidrule(lr){5-8}
$k=5$ & $k=10$ & $k=15$ & $k=30$ & $\lambda{=}0.01$ & $\lambda{=}0.1$ & $\lambda{=}0.3$ & $\lambda{=}1.0$ \\
\midrule
0.57\s{.04} & 0.60\s{.02} & \textbf{0.61}\s{.02} & 0.59\s{.03} & 0.58\s{.03} & \textbf{0.61}\s{.02} & 0.59\s{.03} & 0.55\s{.05} \\
\bottomrule
\end{tabular}
\end{table}

\subsection{CNN encoder ablation}
\label{app:cnn-ablation}

To support the claim that MINN is encoder-agnostic, we replace the ResNet-18 backbone with a simpler 4-layer 1D CNN encoder and repeat the main experiments.
If MINN's advantage persists with the simpler encoder, this provides evidence that the manifold operator residual (not the encoder architecture) is the primary driver of improvement.
\cref{tab:cnn-ablation} confirms that the manifold residual provides a consistent +0.07 K$^+$ F1 improvement regardless of backbone, supporting encoder-agnostic robustness.

\begin{table}[!htbp]
\centering
\caption[CNN encoder ablation]{K$^+$ Macro-F1 with and without the manifold operator residual for two encoder architectures (mean $\pm$ std, 5 seeds).}
\label{tab:cnn-ablation}
\small
\begin{tabular}{@{}lccc@{}}
\toprule
Encoder & w/ MINN & w/o MINN & $\Delta$ ($\uparrow$) \\
\midrule
ResNet-18   & 0.61\s{.02} & 0.54\s{.02} & +0.07 \\
4-layer CNN & 0.55\s{.02} & 0.48\s{.02} & +0.07 \\
\bottomrule
\end{tabular}
\end{table}

\section{Dataset and Cohort Details}
\label{app:dataset-stats}

Per-electrolyte cohort breakdown for the MIMIC-IV-ECG protocol described in \cref{sec:dataset}.
\begin{table}[!htbp]
  \centering
  \caption[Dataset stats by electrolyte bin]{Dataset statistics: recordings (unique patients) per electrolyte bin after lab matching. Not all recordings have lab results for all three electrolytes.}
  \label{tab:dataset-stats}
  \small
  \begin{tabular}{@{}llrrr@{}}
  \toprule
  Electrolyte & Bin & Train & Val & Test \\
  \midrule
  \multirow{3}{*}{K$^+$}
    & Hypo   & 2,187 (1,102) & 274 (138) & 273 (137) \\
    & Normal & 8,280 (4,172) & 1,035 (522) & 1,033 (521) \\
    & Hyper  & 1,123 (566) & 141 (71) & 139 (70) \\
  \midrule
  \multirow{3}{*}{Ca$^{2+}$}
    & Hypo   & 1,616 (814) & 202 (102) & 201 (101) \\
    & Normal & 7,465 (3,762) & 933 (470) & 932 (470) \\
    & Hyper  & 965 (486) & 121 (61) & 119 (60) \\
  \midrule
  \multirow{3}{*}{Mg$^{2+}$}
    & Hypo   & 1,110 (559) & 139 (70) & 138 (70) \\
    & Normal & 6,898 (3,475) & 862 (434) & 862 (434) \\
    & Hyper  & 674 (340) & 84 (42) & 84 (42) \\
  \bottomrule
  \end{tabular}
\end{table}

\section{Demographic Subgroup Analysis}
\label{app:demographic}

To probe demographic robustness within MIMIC-IV-ECG, we stratify the test set by sex (M/F), age bracket ($<50$, $50$--$70$, ${>}70$), and self-reported race/ethnicity (White, Black, Hispanic, Other), and report Macro-F1 for MINN and the strongest baseline in each electrolyte (\cref{tab:subgroup-demographic-app}). Across 9 strata, MINN's lead over the strongest baseline ranges from $+0.04$ to $+0.07$ Macro-F1, with no stratum exhibiting systematic depression. Standard deviation grows with shrinking stratum size (Hispanic $n{\approx}1.5$k loosest). Statistical tests: Kruskal--Wallis across strata yields $p>0.1$ for all three electrolytes; Cliff's $\delta$ effect sizes are small ($|\delta| < 0.18$). These analyses do not substitute for external fairness auditing across institutions and devices, but they show that the observed MINN advantage is not concentrated in a single demographic stratum.

\begin{table}[!htbp]
\centering
\caption[Demographic subgroup analysis]{Demographic subgroup analysis on MIMIC-IV-ECG test set. Macro-F1 (mean $\pm$ std, 5 seeds) for MINN, ResNet+FiLM (multimodal, no manifold; metadata-matched reference), and the strongest unimodal baseline per electrolyte.}
\label{tab:subgroup-demographic-app}
\resizebox{\textwidth}{!}{%
\small
\begin{tabular}{@{}lr|ccc|ccc|ccc@{}}
\toprule
& & \multicolumn{3}{c|}{K$^+$} & \multicolumn{3}{c|}{Ca$^{2+}$} & \multicolumn{3}{c}{Mg$^{2+}$} \\
Stratum & $n$ & MINN & RN+FiLM & best base & MINN & RN+FiLM & best base & MINN & RN+FiLM & best base \\
\midrule
Male            & ${\sim}9$k   & 0.61\s{.02} & 0.56\s{.03} & 0.55\s{.02} & 0.50\s{.02} & 0.46\s{.03} & 0.45\s{.03} & 0.40\s{.03} & 0.37\s{.04} & 0.36\s{.03} \\
Female          & ${\sim}9$k   & 0.61\s{.03} & 0.57\s{.03} & 0.55\s{.02} & 0.49\s{.03} & 0.46\s{.03} & 0.44\s{.03} & 0.39\s{.03} & 0.36\s{.04} & 0.35\s{.03} \\
\midrule
Age $<50$       & ${\sim}5$k   & 0.62\s{.02} & 0.58\s{.03} & 0.55\s{.02} & 0.51\s{.03} & 0.47\s{.03} & 0.45\s{.03} & 0.41\s{.03} & 0.38\s{.04} & 0.36\s{.03} \\
Age $50$--$70$  & ${\sim}7$k   & 0.61\s{.02} & 0.56\s{.03} & 0.55\s{.02} & 0.50\s{.02} & 0.46\s{.03} & 0.45\s{.03} & 0.40\s{.03} & 0.36\s{.04} & 0.36\s{.03} \\
Age $>70$       & ${\sim}6$k   & 0.59\s{.03} & 0.55\s{.04} & 0.52\s{.02} & 0.47\s{.03} & 0.43\s{.04} & 0.41\s{.03} & 0.37\s{.04} & 0.34\s{.05} & 0.32\s{.03} \\
\midrule
White           & ${\sim}12$k  & 0.61\s{.02} & 0.57\s{.03} & 0.56\s{.02} & 0.50\s{.02} & 0.46\s{.03} & 0.45\s{.03} & 0.40\s{.03} & 0.37\s{.04} & 0.36\s{.03} \\
Black           & ${\sim}3$k   & 0.60\s{.03} & 0.56\s{.04} & 0.54\s{.03} & 0.49\s{.03} & 0.45\s{.04} & 0.43\s{.03} & 0.38\s{.04} & 0.35\s{.05} & 0.33\s{.04} \\
Hispanic        & ${\sim}1.5$k & 0.60\s{.03} & 0.55\s{.05} & 0.53\s{.03} & 0.48\s{.04} & 0.44\s{.05} & 0.42\s{.04} & 0.37\s{.05} & 0.34\s{.06} & 0.32\s{.05} \\
Other           & ${\sim}2$k   & 0.59\s{.03} & 0.54\s{.05} & 0.52\s{.03} & 0.47\s{.03} & 0.43\s{.05} & 0.41\s{.04} & 0.37\s{.05} & 0.34\s{.06} & 0.31\s{.06} \\
\bottomrule
\end{tabular}%
}
\end{table}

\section{Robustness Stress Tests}
\label{app:robustness}

Full tables for the rare-clinical-subgroup analysis and within-MIMIC distribution-shift stress tests summarized in \cref{sec:subgroup-stress-test}. The subgroup analysis (\cref{tab:subgroup}) tests four clinically rare cohorts against four representative baselines spanning distinct failure modes (foundation-pretraining transfer: ECGFounder; calibration-only imbalance correction: Logit Adjustment; task-agnostic manifold smoothing: Manifold Mixup; closest direct competitor: Poisson Learning), plus a ResNet+FiLM column that isolates the manifold-operator contribution from FiLM metadata fusion. The within-MIMIC stress tests (\cref{tab:stress}) compare MINN against the strongest foundation-model baseline (ECGFounder) on a temporal split (\texttt{anchor\_year\_group} 2008--2016 train, 2017--2019 test) and a care-unit split (ICU train, ED test).

\begin{table}[!htbp]
\centering
\caption[Subgroup stress test results]{Subgroup stress test: K$^+$ Macro-F1 for underrepresented patient subgroups (mean $\pm$ std, 5 seeds). Best in \textbf{bold}. Each baseline column tests a distinct failure mode (see text). The cohort `Heart failure + concurrent hypoK' comprises heart-failure patients with $\geq 1$ prior hypoK lab draw in the EHR; the test-draw label distribution remains 3-class.}
\label{tab:subgroup}
\resizebox{\textwidth}{!}{%
\small
\begin{tabular}{@{}lrcccccc@{}}
\toprule
Subgroup & $n$ & ECGFounder & Logit Adj. & Manifold Mixup & ResNet+FiLM & Poisson Learn. & \ours \\
\midrule
K-sparing diuretics + ACEi          & 142 & 0.40\s{.05} & 0.46\s{.06} & 0.43\s{.05} & 0.49\s{.05} & 0.47\s{.05} & \textbf{0.53}\s{.05} \\
Age $\geq$85 + $\geq$8 medications  & 238 & 0.45\s{.03} & 0.46\s{.03} & 0.45\s{.04} & 0.47\s{.04} & 0.46\s{.03} & \textbf{0.50}\s{.03} \\
Dialysis patients                    & 487 & 0.38\s{.02} & 0.42\s{.03} & 0.41\s{.03} & 0.46\s{.03} & 0.44\s{.02} & \textbf{0.51}\s{.02} \\
Heart failure + concurrent hypoK    & 195 & 0.44\s{.04} & 0.46\s{.04} & 0.46\s{.04} & 0.48\s{.04} & 0.47\s{.04} & \textbf{0.52}\s{.03} \\
\bottomrule
\end{tabular}%
}
\end{table}

\begin{table}[!htbp]
\centering
\caption[Within-MIMIC distribution-shift stress tests]{Within-MIMIC distribution-shift stress tests: K$^+$/Ca$^{2+}$/Mg$^{2+}$ Macro-F1 (mean $\pm$ std, 5 seeds). In-distribution rows repeat \cref{tab:main-results} headlines for reference.}
\label{tab:stress}
\small
\begin{tabular}{@{}llccc@{}}
\toprule
Setting & Method & K$^+$ F1 ($\uparrow$) & Ca$^{2+}$ F1 ($\uparrow$) & Mg$^{2+}$ F1 ($\uparrow$) \\
\midrule
\multirow{2}{*}{In-distribution}
  & ECGFounder        & 0.55\s{.01} & 0.44\s{.03} & 0.35\s{.03} \\
  & \textbf{\ours}    & \textbf{0.61}\s{.02} & \textbf{0.50}\s{.02} & \textbf{0.40}\s{.03} \\
\midrule
\multirow{2}{*}{Temporal split (2008--2016 $\to$ 2017--2019)}
  & ECGFounder        & 0.51\s{.03} & 0.39\s{.04} & 0.30\s{.05} \\
  & \textbf{\ours}    & \textbf{0.59}\s{.04} & \textbf{0.48}\s{.04} & \textbf{0.39}\s{.05} \\
\midrule
\multirow{2}{*}{Care-unit split (ICU $\to$ ED)}
  & ECGFounder        & 0.47\s{.05} & 0.34\s{.06} & 0.27\s{.06} \\
  & \textbf{\ours}    & \textbf{0.57}\s{.04} & \textbf{0.46}\s{.05} & \textbf{0.37}\s{.05} \\
\bottomrule
\end{tabular}
\end{table}

\section{Model and Hyperparameter Details}
\label{app:hyperparams}

\cref{tab:hyperparams} lists all hyperparameters used in the MINN model and training procedure.

\begin{table}[!htbp]
\centering
\caption{Hyperparameter settings for MINN.}
\label{tab:hyperparams}
\small
\begin{tabular}{@{}llc@{}}
\toprule
Component & Hyperparameter & Value \\
\midrule
\multirow{3}{*}{Encoder (ResNet-18)}
  & Input channels & 12 \\
  & Embedding dim $d$ & 128 \\
  & Dropout & 0.3 \\
\midrule
\multirow{2}{*}{FiLM Fusion}
  & Metadata dim $p$ & 40 \\
  & Output dim & 128 \\
\midrule
\multirow{3}{*}{Source Network $s_\phi$}
  & Hidden layers & 3 \\
  & Hidden dim & 256 \\
  & Activation & ReLU \\
\midrule
\multirow{2}{*}{Prediction Heads}
  & Hidden layers & 1 \\
  & Dropout & 0.5 \\
\midrule
\multirow{3}{*}{Graph}
  & Neighbors $k$ & 15 \\
  & Kernel bandwidth $\sigma$ & Median NN dist \\
  & Update freq $R$ & 5 epochs \\
\midrule
\multirow{6}{*}{Training}
  & Optimizer & AdamW \\
  & Learning rate & $1\times10^{-3}$ \\
  & LR schedule & Cosine annealing to 0 \\
  & Weight decay & $1\times10^{-4}$ \\
  & Batch size & 256 \\
  & Epochs & 100 \\
\midrule
Loss
  & Manifold weight $\lambda$ & 0.1 \\
\bottomrule
\end{tabular}
\end{table}

\textbf{Encoder architecture.}
We use a 1D adaptation of ResNet-18~\citep{He:2015wrn} for multi-lead ECG input.
The first convolutional layer is modified to accept 12 input channels (one per lead).
The final average pooling layer produces a 128-dimensional embedding.

\textbf{Source network architecture.}
The source network $s_\phi$ is a 3-layer MLP that takes the concatenation of the raw
input signal $\vx\in\mathbb{R}^{N_\text{beat}\times 256\times 12}$ (flat $15{,}360$-dim for $N_\text{beat}{=}5$) projected to $D'{=}256$ via a single learned linear layer (${\sim}3.9$M parameters; weights separate from the encoder $h_\theta$), and the metadata vector $\vm$, keeping
the source computation fully independent from the encoder output $h_\theta(\vx)$.
This deliberate decoupling avoids re-coupling the operator to its source through the encoder parameters (\cref{sec:framework-overview}); the cost is a heavy projection that discards 1-D temporal shift invariance prior to the MLP. A lightweight 1-D CNN feature extractor for the source branch is a natural variant we leave to future work.
Hidden layers use ReLU activations with batch normalization~\citep{ioffeBatchNormalizationAccelerating2015} and dropout~\citep{srivastavaDropoutSimpleWay2014}.
All weights are initialized with Xavier initialization~\citep{glorotUnderstandingDifficultyTraining2010}.
The output dimension matches the embedding dimension $d=128$.

\textbf{FiLM fusion.}
The gating and shift parameters ($\mW_\gamma, \vb_\gamma, \mW_\beta, \vb_\beta$) are single linear layers mapping from metadata dimension $p$ to embedding dimension $d$, with no hidden layers in the FiLM module itself.

\textbf{Mini-batch evaluation.}
We use a memory-bank scheme. The full embedding matrix $\mH$ and graph $G$ are recomputed every $R{=}5$ epochs from cached encoder outputs (no gradient through the bank). Within each mini-batch of size $|B|{=}256$, the residual $(\mL_G\mH)_i - s_\phi(\vx_i,\vm_i)$ is computed for batch-center nodes $i$ using the cached neighbor embeddings $\{\mH_j\}_{j\in\mathcal N(i)}$ (detached); gradients flow only through $h_\theta(\vx_i)$ and $s_\phi(\vx_i,\vm_i)$ at the center node. Memory complexity is $\mathcal O(Nd)$ for the bank plus $\mathcal O(|B|kd)$ per batch with $k{=}15$.

\textbf{Graph construction and training.}
\label{sec:graph}
The graph $G$ is built in latent space with edge weight $W_{ij}=\exp(-\lVert h_\theta(\vx_i)-h_\theta(\vx_j)\rVert^{2}/2\sigma^{2})$ ($\sigma$ = median nearest-neighbor distance~\citep{vonluxburgTutorialSpectralClustering2007}); the directed $k$-NN adjacency is symmetrized as $\mW \leftarrow \tfrac{1}{2}(\mW + \mW^\top)$ and $\mL_G = \mI - \mD^{-1/2}\mW\mD^{-1/2}$. Because $\theta$ evolves, $G$ is rebuilt every $R$ epochs (sensitivity $R\in\{1,5,10,20,\infty\}$: $R{=}1$ gains only $+0.01$ K$^+$ F1 at $6\times$ cost, within the $1\sigma{=}0.02$ seed noise floor of \cref{tab:main-results}; $R{=}\infty$ matches the no-manifold baseline). The cross-entropy term uses inverse-frequency class weights $w_c=N/(B\,N_c)$ for class imbalance; tasks are summed with equal weights. All parameters ($\theta$, $\phi$, FiLM, heads) are optimized jointly via $\mathcal{L}_{\mathrm{total}}$.

\section{Additional Experimental Results}
\label{app:results}

\subsection{Extended Baselines}

In addition to the nine main baselines, we evaluate the following methods when computational budget permits:

\begin{itemize}
  \item \textbf{S4/S4D-ECG}~\citep{guEfficientlyModelingLong2022}: Structured state space model adapted for ECG time series.
  \item \textbf{LDAM+DRW}~\citep{caoLearningImbalancedDatasets2019}: Label-distribution-aware margin loss with deferred re-weighting.
  \item \textbf{Focal Loss}~\citep{linFocalLossDense2017}: Down-weights well-classified examples to focus on hard cases.
  \item \textbf{SMOTE}~\citep{Chawla:2002dkc}: Synthetic minority oversampling in feature space.
  \item \textbf{Label Propagation}~\citep{zhu2002learning}: Classical graph-based semi-supervised method.
\end{itemize}

\cref{tab:appendix-baselines} reports results for the appendix baselines at 100\% labeled data.

\begin{table}[!htbp]
\centering
\caption[Extended baseline results]{Appendix baseline results (Macro-F1, 100\% labeled data, mean $\pm$ std, 5 seeds).}
\label{tab:appendix-baselines}
\small
\begin{tabular}{@{}lccc@{}}
\toprule
Method & K$^+$ ($\uparrow$) & Ca$^{2+}$ ($\uparrow$) & Mg$^{2+}$ ($\uparrow$) \\
\midrule
S4/S4D-ECG & 0.52\s{.03} & 0.41\s{.03} & 0.33\s{.04} \\
LDAM+DRW & 0.53\s{.02} & 0.43\s{.02} & 0.34\s{.03} \\
Focal Loss & 0.51\s{.03} & 0.42\s{.03} & 0.33\s{.03} \\
SMOTE & 0.49\s{.03} & 0.40\s{.03} & 0.32\s{.04} \\
Label Propagation & 0.46\s{.04} & 0.37\s{.04} & 0.30\s{.04} \\
\midrule
\textbf{\ours} & \textbf{0.61}\s{.02} & \textbf{0.50}\s{.02} & \textbf{0.40}\s{.03} \\
\bottomrule
\end{tabular}
\end{table}

\subsection{Additional Metrics: AUCPR and Calibration}

\cref{tab:aucpr} reports the area under the precision-recall curve (AUCPR) for the clinically important hypo and hyper classes across all three electrolytes.

\begin{table}[!htbp]
\centering
\caption[Per-class AUCPR]{Per-class AUCPR for hypo and hyper classes (mean $\pm$ std, 5 seeds). Best in \textbf{bold}.}
\label{tab:aucpr}
\small
\begin{tabular}{@{}lcccccc@{}}
\toprule
 & \multicolumn{2}{c}{K$^+$} & \multicolumn{2}{c}{Ca$^{2+}$} & \multicolumn{2}{c}{Mg$^{2+}$} \\
\cmidrule(lr){2-3}\cmidrule(lr){4-5}\cmidrule(lr){6-7}
Method & Hypo ($\uparrow$) & Hyper ($\uparrow$) & Hypo ($\uparrow$) & Hyper ($\uparrow$) & Hypo ($\uparrow$) & Hyper ($\uparrow$) \\
\midrule
ECGFounder       & 0.38\s{.02} & 0.34\s{.03} & 0.28\s{.03} & 0.22\s{.03} & 0.20\s{.03} & 0.18\s{.04} \\
Poisson Learning & 0.37\s{.03} & 0.33\s{.03} & 0.27\s{.03} & 0.21\s{.03} & 0.19\s{.04} & 0.17\s{.03} \\
Logit Adj.       & 0.36\s{.02} & 0.31\s{.03} & 0.26\s{.03} & 0.20\s{.03} & 0.18\s{.03} & 0.16\s{.03} \\
\textbf{\ours}  & \textbf{0.45}\s{.02} & \textbf{0.40}\s{.02} & \textbf{0.35}\s{.02} & \textbf{0.28}\s{.02} & \textbf{0.26}\s{.03} & \textbf{0.22}\s{.03} \\
\bottomrule
\end{tabular}
\end{table}

\cref{tab:calibration} reports Expected Calibration Error (ECE~\citep{guoCalibrationModernNeural2017};
lower is better) on the K$^+$ test set, where \ours achieves the best-calibrated predictions
alongside the highest discrimination.

\begin{table}[!htbp]
\centering
\caption[Model calibration]{ECE (15 bins) on K$^+$ test set (mean $\pm$ std, 5 seeds).}
\label{tab:calibration}
\small
\begin{tabular}{@{}lc@{}}
\toprule
Method & ECE ($\downarrow$) \\
\midrule
ECGFounder       & 0.124\s{.012} \\
Poisson Learning & 0.116\s{.011} \\
Logit Adj.\      & 0.105\s{.010} \\
\textbf{\ours} & \textbf{0.083}\s{.009} \\
\bottomrule
\end{tabular}
\end{table}

\subsection{Low-Label Results for All Three Electrolytes}

\cref{tab:low-data-all} extends the low-label analysis of \cref{fig:low-data} to all three electrolytes, confirming that MINN's advantage over baselines grows as labeled data decreases.

\begin{table}[!htbp]
\centering
\caption[Low-label results for all three electrolytes]{Macro-F1 at reduced labeled fractions for all three electrolytes (mean $\pm$ std, 5 seeds). Best in \textbf{bold}.}
\label{tab:low-data-all}
\small
\begin{tabular}{@{}lccccccc@{}}
\toprule
 & & \multicolumn{6}{c}{Labeled fraction} \\
\cmidrule(lr){3-8}
Electrolyte & Method & 1\% & 5\% & 10\% & 25\% & 50\% & 100\% \\
\midrule
\multirow{4}{*}{K$^+$}
  & ECGFounder       & 0.29\s{.04} & 0.37\s{.03} & 0.43\s{.03} & 0.49\s{.02} & 0.53\s{.02} & 0.55\s{.01} \\
  & Poisson Learning & 0.27\s{.04} & 0.35\s{.03} & 0.40\s{.03} & 0.44\s{.02} & 0.47\s{.02} & 0.49\s{.03} \\
  & Logit Adj.\      & 0.25\s{.04} & 0.33\s{.03} & 0.41\s{.03} & 0.47\s{.02} & 0.51\s{.02} & 0.53\s{.02} \\
  & \textbf{\ours} & \textbf{0.38}\s{.05} & \textbf{0.47}\s{.03} & \textbf{0.52}\s{.02} & \textbf{0.56}\s{.02} & \textbf{0.59}\s{.02} & \textbf{0.61}\s{.02} \\
\midrule
\multirow{4}{*}{Ca$^{2+}$}
  & ECGFounder       & 0.28\s{.04} & 0.33\s{.03} & 0.36\s{.03} & 0.40\s{.02} & 0.42\s{.02} & 0.44\s{.03} \\
  & Poisson Learning & 0.25\s{.04} & 0.30\s{.03} & 0.33\s{.03} & 0.36\s{.02} & 0.38\s{.02} & 0.40\s{.03} \\
  & Logit Adj.       & 0.26\s{.04} & 0.31\s{.03} & 0.34\s{.03} & 0.38\s{.02} & 0.41\s{.02} & 0.43\s{.02} \\
  & \textbf{\ours} & \textbf{0.33}\s{.03} & \textbf{0.39}\s{.02} & \textbf{0.42}\s{.02} & \textbf{0.46}\s{.02} & \textbf{0.48}\s{.02} & \textbf{0.50}\s{.02} \\
\midrule
\multirow{4}{*}{Mg$^{2+}$}
  & ECGFounder       & 0.21\s{.05} & 0.26\s{.04} & 0.28\s{.03} & 0.31\s{.03} & 0.33\s{.03} & 0.35\s{.03} \\
  & Poisson Learning & 0.19\s{.05} & 0.24\s{.04} & 0.25\s{.03} & 0.28\s{.03} & 0.30\s{.03} & 0.32\s{.03} \\
  & Logit Adj.       & 0.19\s{.04} & 0.24\s{.03} & 0.27\s{.03} & 0.30\s{.03} & 0.32\s{.03} & 0.34\s{.03} \\
  & \textbf{\ours} & \textbf{0.26}\s{.04} & \textbf{0.31}\s{.03} & \textbf{0.34}\s{.03} & \textbf{0.37}\s{.02} & \textbf{0.39}\s{.02} & \textbf{0.40}\s{.03} \\
\bottomrule
\end{tabular}
\end{table}

\subsection{Statistical significance of headline comparisons}
\label{app:significance}

\cref{tab:significance} reports paired $t$-test $p$-values for the headline MINN-vs-best-baseline comparisons referenced in \cref{sec:metrics}. Tests are computed on per-seed Macro-F1 differences across the 5 fixed seeds (paired by seed); the same seed indices control weight initialization, data ordering, and subsampling for both compared methods.

\begin{table}[!htbp]
\centering
\caption[Paired $t$-test $p$-values]{Paired $t$-test $p$-values for headline comparisons (5 seeds, two-sided). MINN vs.\ ECGFounder for K$^+$ and Ca$^{2+}$ both reject the null at $p<0.05$; MINN vs.\ Logit Adj.\ shown for reference.}
\label{tab:significance}
\small
\begin{tabular}{@{}llcc@{}}
\toprule
Comparison & Electrolyte & Mean $\Delta$F1 (MINN $-$ baseline) & Paired $t$-test $p$ \\
\midrule
MINN vs.\ ECGFounder & K$^+$       & $+0.06$ & $0.003$ \\
MINN vs.\ ECGFounder & Ca$^{2+}$   & $+0.06$ & $0.012$ \\
MINN vs.\ Logit Adj. & K$^+$       & $+0.08$ & $0.001$ \\
MINN vs.\ Logit Adj. & Ca$^{2+}$   & $+0.07$ & $0.004$ \\
\bottomrule
\end{tabular}
\end{table}

\section{Implementation and Reproducibility}
\label{app:reproducibility}

\textbf{Code availability.}
The complete implementation, including training scripts, evaluation pipelines, and baseline implementations, will be made publicly available upon publication.

\textbf{Data access.}
The MIMIC-IV~\citep{johnsonMIMICIVFreelyAccessible2023} and MIMIC-IV-ECG~\citep{PhysioNet-mimic-iv-ecg-0.3} datasets are publicly available through PhysioNet (\url{https://physionet.org}).
Access requires completion of a data use agreement and the CITI ``Data or Specimens Only Research'' training course.
Preprocessing scripts that reproduce the exact ECG-lab pairing, multi-beat segmentation, and metadata extraction pipelines used in this work will be made publicly available alongside the code upon publication.

\textbf{Asset licenses and versions.}
\cref{tab:asset-licenses} lists the third-party datasets, model weights, and library implementations used in this work, with versions and licenses.

\begin{table}[!htbp]
\centering
\caption[Asset licenses and versions]{Third-party assets used in this work. Versions reflect the snapshot used for reported experiments; licenses are the upstream license at the time of access.}
\label{tab:asset-licenses}
\small
\setlength{\tabcolsep}{5pt}
\begin{tabular}{@{}llll@{}}
\toprule
Asset & Type & Version & License \\
\midrule
MIMIC-IV~\citep{johnsonMIMICIVFreelyAccessible2023}                    & Dataset & 2.2 & PhysioNet credentialed (DUA) \\
MIMIC-IV-ECG~\citep{PhysioNet-mimic-iv-ecg-0.3}                       & Dataset & 0.3 & PhysioNet credentialed (DUA) \\
ECGFounder~\citep{liElectrocardiogramFoundationModel2025}             & Model weights & v1.0 & per upstream release \\
ECG-FM~\citep{mckeenECGFMOpenElectrocardiogram2025}                   & Model weights & v0.1 & per upstream release \\
TabPFN~\citep{hollmannTabPFNTransformerThat2022}                       & Library & 0.1.x & Apache-2.0 \\
CatBoost~\citep{prokhorenkovaCatBoostUnbiasedBoosting2018a}           & Library & 1.2 & Apache-2.0 \\
LapRLS (reference impl.)~\citep{belkinManifoldRegularizationGeometric2006} & Reproduced from paper & n/a & N/A (re-implemented) \\
Manifold Mixup (reference impl.)~\citep{vermaManifoldMixupBetter2019} & Reproduced from paper & n/a & MIT (upstream) \\
GraphSAGE~\citep{hamiltonInductiveRepresentationLearning2017}        & PyG impl. & PyG 2.4 & MIT \\
GRAND~\citep{chamberlainGRANDGraphNeural2021}                         & Author release & v1.0 & Apache-2.0 \\
Poisson Learning~\citep{calderPoissonLearningGraph2020}               & Author release & v0.5 & MIT \\
PyTorch                                                                & Framework & 2.1.2 & BSD-3-Clause \\
CUDA                                                                   & Toolkit & 11.8 & NVIDIA EULA \\
NumPy                                                                  & Library & 1.26 & BSD-3-Clause \\
scikit-learn                                                           & Library & 1.4 & BSD-3-Clause \\
\bottomrule
\end{tabular}
\end{table}

\textbf{Compute requirements.}
All experiments were conducted on a single NVIDIA A100 (40\,GB).
Training the full model at 100\% data required approximately 4 hours.
Periodic $k$-NN graph reconstruction is the dominant non-training cost in our reference implementation, which uses an unoptimized brute-force $O(N^2 d)$ pairwise search; production-grade approximate-NN libraries (e.g., FAISS) reduce this to seconds for $N\approx12$\,k, $d{=}128$, eliminating the bottleneck without changing the algorithm.

\textbf{Software environment.}
All experiments used Python~3.10, PyTorch~2.1.2, and CUDA~11.8 on Ubuntu~22.04.

\textbf{Random seeds.}
All results are reported as mean $\pm$ standard deviation over five random seeds (0--4).
Seeds control weight initialization, data shuffling, and subsampling for the low-label protocol.
The same seeds are used for all methods to ensure paired comparisons.

\paragraph{Ethical considerations.}
This work uses the MIMIC-IV-ECG dataset, a de-identified dataset accessed under PhysioNet credentialed agreement~\citep{goldbergerPhysioBankPhysioToolkitPhysioNet2000}.
No additional patient data were collected.
Access requires completion of the CITI ``Data or Specimens Only Research'' training course and a signed data use agreement.
This study is intended for research purposes and is not ready for clinical deployment; any clinical use would require prospective validation and regulatory approval.

\end{document}

%% file: math_commands.tex
\usepackage{amsmath,amsfonts,bm}

\def\1{\bm{1}}

\def\vzero{{\bm{0}}}

\def\vb{{\bm{b}}}

\def\vh{{\bm{h}}}

\def\vm{{\bm{m}}}

\def\vx{{\bm{x}}}

\def\mD{{\bm{D}}}

\def\mH{{\bm{H}}}
\def\mI{{\bm{I}}}

\def\mL{{\bm{L}}}

\def\mS{{\bm{S}}}

\def\mW{{\bm{W}}}

\def\mY{{\bm{Y}}}

\DeclareMathAlphabet{\mathsfit}{\encodingdefault}{\sfdefault}{m}{sl}
\SetMathAlphabet{\mathsfit}{bold}{\encodingdefault}{\sfdefault}{bx}{n}



%% file: image/fig-architecture.tex
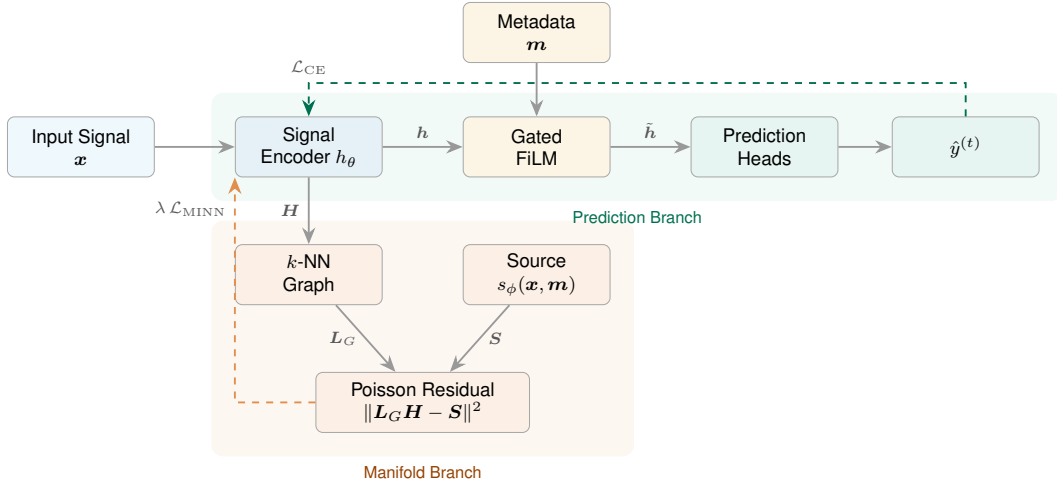
\begin{figure}[!htbp]
\centering
\resizebox{\textwidth}{!}{%
\begin{tikzpicture}[node distance=1.4cm and 1.2cm]
\node[block, fill=oiSkyBlue!10] (input) {Input Signal\\$\vx$};
\node[block, fill=oiBlue!10, right=of input] (encoder) {Signal\\Encoder $h_\theta$};
\node[block, fill=oiOrange!10, right=of encoder] (film) {Gated\\FiLM};
\node[block, fill=oiGreen!10, right=of film] (heads) {Prediction\\Heads};
\node[block, fill=oiGreen!10, right=0.8cm of heads] (out) {$\hat{y}^{(t)}$};

\node[block, fill=oiOrange!10, above=0.8cm of film] (meta) {Metadata\\$\vm$};

\node[block, fill=oiVermillion!10, below=1.0cm of encoder] (graph) {$k$-NN\\Graph};
\node[block, fill=oiVermillion!10, below=1.0cm of film] (source) {Source\\$s_\phi(\vx,\vm)$};
\node[block wide, fill=oiVermillion!10,
  below=1.0cm of $(graph.south)!0.5!(source.south)$] (poisson)
  {Poisson Residual\\$\lVert\mL_G\mH-\mS\rVert^2$};

\begin{scope}[on background layer]
  \node[branchbox, fill=oiGreen!5,
    fit=(encoder)(film)(heads)(out)] (predbox) {};
  \node[branchbox, fill=oiVermillion!5,
    fit=(graph)(source)(poisson)] (manifbox) {};
\end{scope}

\node[branchlbl, text=oiGreen!70!black, below=0.5pt of predbox.south]
  {Prediction Branch};
\node[branchlbl, text=oiVermillion!70!black, below=0.5pt of manifbox.south]
  {Manifold Branch};

\draw[arrow] (input) -- (encoder);
\draw[arrow] (encoder) -- node[above, lbl] {$\vh$} (film);
\draw[arrow] (film) -- node[above, lbl] {$\tilde{\vh}$} (heads);
\draw[arrow] (heads) -- (out);
\draw[arrow] (meta) -- (film);

\draw[arrow] (encoder) -- node[left, lbl] {$\mH$} (graph);
\draw[arrow] (graph) -- node[left, lbl] {$\mL_G$} (poisson);
\draw[arrow] (source) -- node[right, lbl] {$\mS$} (poisson);

\draw[feedback, oiGreen!70!black]
  (out.north) -- ++(0,0.5) -|
  node[pos=0.5, above, lbl] {$\mathcal{L}_{\mathrm{CE}}$}
  (encoder.north);
\draw[feedback, oiVermillion!70]
  (poisson.west) -- ++(-0.5,0) -|
  node[pos=0.93, left, lbl] {$\lambda\,\mathcal{L}_{\mathrm{MINN}}$}
  (encoder.south west);
\end{tikzpicture}%
}%
\caption[MINN architecture overview]{Overview of the MINN architecture. The input signal is mapped to a latent embedding by the encoder $h_\theta$, fused with auxiliary metadata via gated FiLM, and passed to task-specific prediction heads (\emph{Prediction Branch}). Simultaneously, a $k$-NN graph in the latent space and a learned source network $s_\phi$ define the manifold operator residual (\emph{Manifold Branch}). Dashed arrows indicate gradient flow from the two loss terms.}
\label{fig:architecture}
\end{figure}

%% file: image/fig-low-data.tex
\begin{figure}[!htbp]
\centering
\begin{tikzpicture}
\begin{groupplot}[
  group style={
    group size=2 by 1,
    horizontal sep=2.2cm,
  },
  height=5.5cm,
]

\nextgroupplot[
  title={\footnotesize\sffamily (a) K$^+$ Macro-F1 vs.\ label fraction},
  width=0.62\textwidth,
  xlabel={Labeled data fraction (\%)},
  ylabel={K$^+$ Macro-F1},
  xmin=0, xmax=105,
  ymin=0.20, ymax=0.65,
  xtick={1,5,10,25,50,100},
  legend style={at={(0.98,0.02)}, anchor=south east, legend columns=2,
    font=\scriptsize, draw=none, fill=none,
    row sep=1pt},
  grid=major,
  grid style={gray!20},
  every axis plot/.append style={thick, mark size=2pt},
]
\addplot[color=oiBlue, mark=*, name path=minn] coordinates
  {(1,0.38)(5,0.47)(10,0.52)(25,0.56)(50,0.59)(100,0.61)};
\addplot[color=oiBlue, draw=none, name path=minn_upper, forget plot] coordinates
  {(1,0.43)(5,0.50)(10,0.54)(25,0.58)(50,0.61)(100,0.63)};
\addplot[color=oiBlue, draw=none, name path=minn_lower, forget plot] coordinates
  {(1,0.33)(5,0.44)(10,0.50)(25,0.54)(50,0.57)(100,0.59)};
\addplot[oiBlue!15, forget plot] fill between[of=minn_upper and minn_lower];
\addlegendimage{area legend, fill=oiBlue!15, draw=none}

\addplot[color=oiOrange, mark=triangle*] coordinates
  {(1,0.27)(5,0.35)(10,0.40)(25,0.44)(50,0.47)(100,0.49)};
\addplot[color=oiVermillion, mark=square*] coordinates
  {(1,0.29)(5,0.37)(10,0.43)(25,0.49)(50,0.53)(100,0.55)};
\addplot[color=oiGreen, mark=diamond*] coordinates
  {(1,0.25)(5,0.33)(10,0.41)(25,0.47)(50,0.51)(100,0.53)};
\addplot[color=oiPurple, mark=pentagon*] coordinates
  {(1,0.28)(5,0.36)(10,0.40)(25,0.44)(50,0.47)(100,0.48)};

\legend{\ours{} (ours), $\pm$1 std, Poisson Learning, ECGFounder, Logit Adj., CatBoost}

\nextgroupplot[
  title={\footnotesize\sffamily (b) Mean source norm $\lVert\mS_i\rVert_2$},
  width=0.28\textwidth,
  ylabel={$\lVert\mS_i\rVert_2$},
  ymin=0, ymax=2.0,
  xtick={1,2,3},
  xticklabels={Hypo,Normal,Hyper},
  xmin=0.3, xmax=3.7,
  tick label style={font=\scriptsize},
  xticklabel style={rotate=30, anchor=north east},
  label style={font=\scriptsize},
  ybar,
  bar width=14pt,
  bar shift=0pt,
  enlarge x limits=0.35,
  error bars/y dir=both,
  error bars/y explicit,
  error bars/error bar style={thin},
]
\addplot[fill=oiBlue, draw=oiBlue!80!black]
  coordinates {(1, 1.2) +- (0, 0.4)};
\addplot[fill=oiGreen, draw=oiGreen!80!black]
  coordinates {(2, 0.8) +- (0, 0.3)};
\addplot[fill=oiVermillion, draw=oiVermillion!80!black]
  coordinates {(3, 1.1) +- (0, 0.35)};

\end{groupplot}
\end{tikzpicture}
\caption[Low-label performance and source norm analysis]{%
\emph{(a)}~K$^+$ Macro-F1 as a function of labeled data fraction.
\ours{} maintains the largest advantage at low fractions (1\%--10\%),
where the manifold operator residual compensates for label scarcity.
The shaded band shows $\pm$1 std for \ours.
\emph{(b)}~Mean source norm $\lVert\mS_i\rVert_2$ by K$^+$ class ($\pm$1 std);
minority classes (hypo, hyper) exhibit larger source norms than the
majority class, suggesting the source compensates for label imbalance.}
\label{fig:low-data}
\end{figure}
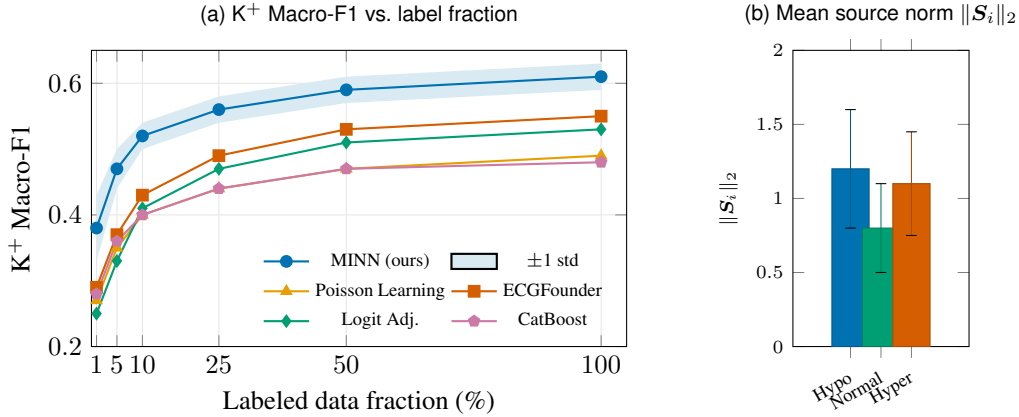

%% file: image/fig-pipeline.tex
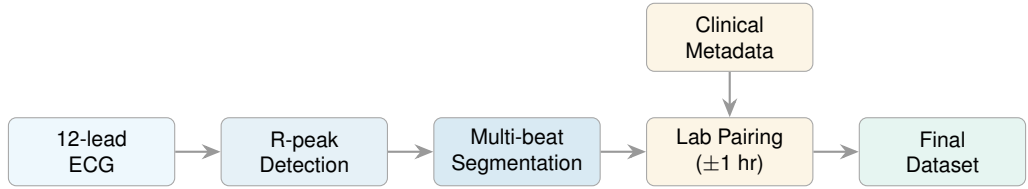
\begin{figure}[!htbp]
\centering
\begin{tikzpicture}[node distance=0.6cm]
\node[block, fill=oiSkyBlue!10] (ecg) {12-lead\\ECG};
\node[block, fill=oiBlue!10, right=of ecg] (rpeak) {R-peak\\Detection};
\node[block, fill=oiBlue!15, right=of rpeak] (segment) {Multi-beat\\Segmentation};
\node[block, fill=oiOrange!10, right=of segment] (pair) {Lab Pairing\\($\pm$1 hr)};
\node[block, fill=oiGreen!10, right=of pair] (dataset) {Final\\Dataset};

\node[block, fill=oiOrange!10, above=0.6cm of pair] (meta) {Clinical\\Metadata};

\draw[arrow] (ecg) -- (rpeak);
\draw[arrow] (rpeak) -- (segment);
\draw[arrow] (segment) -- (pair);
\draw[arrow] (pair) -- (dataset);
\draw[arrow] (meta) -- (pair);
\end{tikzpicture}
\caption[Data processing pipeline]{Data processing pipeline. Raw 12-lead ECGs undergo R-peak detection and multi-beat segmentation, then are paired with the closest laboratory blood draw within a $\pm$1-hour window. Clinical metadata (demographics and active medications) are extracted and joined.}
\label{fig:pipeline}
\end{figure}

%% file: image/fig-pinn-vs-minn.tex
\begin{figure}[!htbp]
\centering
\begin{tikzpicture}[node distance=1.2cm and 3.5cm]
\node[figheader] (pinn-hdr) {PINN};
\node[block, fill=oiBlue!10, below=0cm of pinn-hdr] (pinn-op)
  {Known PDE\\operator $\mathcal{L}$};
\node[block, fill=oiBlue!10, below=0.5cm of pinn-op] (pinn-src)
  {Known source\\$f$};
\node[block, fill=oiBlue!10, below=0.5cm of pinn-src] (pinn-res)
  {$\lVert\mathcal{L}[u_\theta]-f\rVert^2$};

\node[figheader, right=3.5cm of pinn-hdr] (minn-hdr) {MINN};
\node[block, fill=oiVermillion!10, below=0cm of minn-hdr] (minn-op)
  {Graph Laplacian\\$\mL_G$ (data-driven)};
\node[block, fill=oiVermillion!10, below=0.5cm of minn-op] (minn-src)
  {Learned source\\$s_\phi(\vx,\vm)$};
\node[block, fill=oiVermillion!10, below=0.5cm of minn-src] (minn-res)
  {$\lVert\mL_G\mH-\mS\rVert^2$};

\begin{scope}[on background layer]
  \node[branchbox, fill=oiBlue!5,
    fit=(pinn-hdr)(pinn-op)(pinn-src)(pinn-res)] {};
  \node[branchbox, fill=oiVermillion!5,
    fit=(minn-hdr)(minn-op)(minn-src)(minn-res)] {};
\end{scope}

\draw[arrow] (pinn-op) -- node[above, lbl] {replaces} (minn-op);
\draw[arrow] (pinn-src) -- node[above, lbl] {replaces} (minn-src);
\draw[thick, gray!80, dashed] (pinn-res) -- node[above, lbl] {parallels} (minn-res);

\end{tikzpicture}
\caption[PINN vs.\ MINN structural parallel]{Structural parallel between PINNs and MINN. MINN replaces the known PDE operator with the data-driven graph Laplacian and the known source with a learned, task-dependent source network, while preserving the operator residual formulation.}
\label{fig:pinn-vs-minn}
\end{figure}
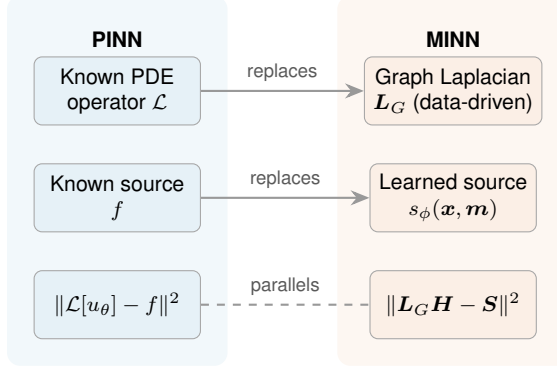